\documentclass[review]{elsarticle}
\usepackage[utf8x]{inputenc}
\usepackage{lineno,hyperref}
\usepackage{graphicx}
\usepackage{subfigure}
\usepackage{amssymb}
\usepackage{amsmath}
\usepackage{threeparttable}
\usepackage{booktabs}
\usepackage{algorithm}
\usepackage{verbatim}
\usepackage{subfigure}
\usepackage{float}

\usepackage{xparse}
\usepackage{wrapfig}
\usepackage{xkeyval} 
\newcommand{\xucomment}[1]{}
\newcommand{\baicomment}[1]{}

\modulolinenumbers[5]

\journal{Neural Networks}

\newwrite\authorbibfile%
\AtBeginDocument{%
  \immediate\openout\authorbibfile=\jobname.aub%
}%

\AtEndDocument{%
\immediate\closeout\authorbibfile
\InputIfFileExists{\jobname.aub}{}{}
}%

\makeatletter

\define@key{authorbib}{scale}[1]{%
\def\AuthorbibKVMacroScale{#1}%
}

\define@key{authorbib}{wraplines}[10]{%
\def\AuthorbibKVMacroWraplines{#1}%
}

\define@key{authorbib}{imagewidth}[4cm]{%
\def\AuthorbibKVMacroImagewidth{#1}%
}

\define@key{authorbib}{overhang}[10pt]{%
\def\AuthorbibKVMacroOverhang{#1}%
}

\define@key{authorbib}{imagepos}[l]{%
\def\AuthorbibKVMacroImagepos{#1}%
}

\makeatother
\presetkeys{authorbib}{imagepos=l,imagewidth=4cm,wraplines=15,overhang=20pt}{}

\newlength{\AuthorbibTopSkip}
\newlength{\AuthorbibBottomSkip}
\NewDocumentCommand{\authorbibliography}{+o+m+m+m}{%
  \IfNoValueTF{#1}{%
  }{%
    \setkeys{authorbib}{#1}%
    \immediate\write\authorbibfile{%
      \string\begin{wrapfigure}[\AuthorbibKVMacroWraplines]{\AuthorbibKVMacroImagepos}[\AuthorbibKVMacroOverhang]{\AuthorbibKVMacroImagewidth}^^J
        \string\includegraphics[scale=\AuthorbibKVMacroScale]{#2}^^J
        \string\end{wrapfigure}^^J
    }%
  }%
  \IfNoValueTF{#3}{%
    \typeout{Warning: No author name}%
  }{%
    \immediate\write\authorbibfile{%
      \unexpanded{\vspace{\AuthorbibTopSkip}}^^J
      \string\noindent\relax
      \unexpanded{\textbf{#3}\par}^^J
      \string\noindent\relax
      \unexpanded{#4}^^J%
      \unexpanded{\vspace{\AuthorbibBottomSkip}}^^J
      }%
  }%
}%

\begin{document}

\begin{frontmatter}

\title{Structured Pruning of Recurrent Neural Networks through Neuron Selection}

\author[1]{Liangjian Wen}
\author[1]{Xuanyang Zhang}
\author[2]{Haoli Bai}
\author[1,3]{Zenglin Xu\corref{mycorrespondingauthor}}

\ead{zlxu@uestc.edu.cn}
\address[1]{SMILE Lab, School of Computer Science and Engineering \\ University of Electronic Science and Technology of China\\ Chengdu 610031, China.}
\address[2]{Department of Computer Science and Engineering\\ The Chinese University of Hong Kong\\ Shatin NT 999077, Hong Kong SAR}
\address[3]{Center of Artificial Intelligence \\ Peng Cheng Lab\\ Shenzhen, Guangdong, China}
\cortext[mycorrespondingauthor]{Corresponding author}


%
%

\begin{abstract}
Recurrent  neural  networks (RNNs) have recently achieved remarkable successes in a number of applications. 
However, the huge sizes and computational burden of these models make it difficult for their deployment on edge devices. 
A practically effective approach is to reduce the overall storage  and computation costs of RNNs by network pruning techniques. Despite their successful applications, those pruning methods based on Lasso either produce irregular sparse patterns in weight matrices, which is not helpful in practical speedup.
To address these issues, we propose a structured pruning method through neuron selection which can remove  the independent neuron of RNNs.
More specifically, we introduce two sets of binary random variables, which can be interpreted as gates or switches to the input neurons and the hidden neurons, respectively. We demonstrate that the corresponding optimization problem can be addressed by minimizing the $L_0$ norm of the weight matrix. 
Finally, experimental results on language modeling and machine reading comprehension tasks have indicated the advantages of the proposed method in comparison with state-of-the-art pruning competitors. 
In particular,  nearly 20$\times$ practical speedup during inference was achieved without losing performance for the language model on the Penn TreeBank dataset, indicating the promising performance of the proposed method.
\end{abstract}

\end{frontmatter}
\section{Introduction}
 Recurrent neural networks (RNNs) have recently achieved remarkable successes in multiple fields such as image captioning~\cite{vinyals2016show,anderson2018bottom}, action recognition~\cite{YeWLCZCX18,Pan2018CompressingRN}, music segmentation~\cite{LiuHBDBX18}, question answering~\cite{SAGARA2014201,MAO20183}, machine translation~\cite{DBLP:journals/corr/BahdanauCB14,DBLP:conf/emnlp/LuongPM15,YANG2018146}, and language modelling~\cite{byeon2015scene,sutskever2014sequence,LiuHBDBX18}.  
 These successes heavily rely on huge models trained on large datasets, especially for those RNN variants such as Long Short Term Memory (LSTM) networks~\cite{hochreiter1997long} and Gated Recurrent Unit (GRU) networks \cite{cho2014learning}.
 With the increasing popularity of edge computing, a recent trend is to deploy these models onto end devices so as to allow off-line reasoning and inference. 
 However, these models are generally of huge sizes and bring expensive computation and storage costs during inference, which makes the deployment difficult for those devices with limited resources. In order to reduce the overall computation and storage costs of these models, model compression on recurrent neural networks has been widely concerned.

Network pruning is one of the prominent approaches to tackle the compression of RNNs. \cite{narang2017exploring} presents a connection pruning method to compress RNNs efficiently. 
However, the obtained weight matrix via connection pruning has random and unstructured sparsity. 
Such unstructured sparse formats are unfriendly for efficient computation in modern hardware systems~\cite{lebedev2016fast,ZhaoWWBVYTX17} due to irregular memory access in modern processors.
Previous studies~\cite{DBLP:conf/nips/WenWWCL16,Wen2017LearningIS} have shown that speedup obtained with random sparse matrix multiplication on various hardware platforms is lower than expected. 
For example, varying the sparsity level in weight matrices of AlexNet in the range of 67.6\%, 92.4\%, 94.3\%, 96.6\%, and 97.2\%, the speedup ratio was 0.25$\times$, 0.52$\times$, 1.36$\times$, 1.04$\times$, and 1.38$\times$, respectively.  
A practical remedy to this problem is structured pruning where 
pruning individual neurons can directly trim weight matrix size such that structured sparse matrix multiplication efficiently utilizes the hardware resources.

Due to the promising properties of structured pruning, the structured pruning on deep neuron networks(DNNs)  has been widely explored~\cite{DBLP:conf/nips/ZhuangTZLGWHZ18,DBLP:conf/iccv/HeZS17,ding2019centripetal,he2019filter}. However, compared with the structured pruning on DNNs, there is a vital challenge originated from recurrent structure of RNNs, which is shared across all the time steps in a sequence. Structured pruning methods used in DNNs cannot be directly applied to RNNs. The reason is that independently removing the links can result in a mismatch of feature dimensions and then induce invalid recurrent units. 
In contrast, this problem does not exist in DNNs, where neurons  can be independently removed without violating the usability of the final network structure.
Accordingly group sparsity~\cite{Louizos2017LearningSN} is difficult to be applied in RNNs.

To address this issue, we explore a new type of method along the line of structured pruning of RNNs through neuron selection. In detail,
we introduce two sets of binary random variables, which can be interpreted as gates to the neurons, to indicate the presence of the input neurons and the hidden neurons, respectively. 
The two sets of binary random variables are then used to generate sparse masks for the weight matrix.
More specifically, the presence of the matrix entry $w_{ij}$ depends on both the presence of the $i$-th input unit and the $j$-th output unit, while the value of $w_{ij}$ indicates the strength of the connection if $w_{ij}\ne 0$. 
However, the optimization of these variables is computationally intractable due to the nature of $2^{|h|}$ possible states of binary gate variable vector $h$.
We then develop an efficient $L_0$ inference algorithm for inferring the binary gate variables, motivated from the work of pruning DNN weights\cite{DBLP:conf/cvpr/SrinivasSB17, Louizos2017LearningSN}. 

While previous efforts on structured pruning of RNNs resort to the group lasso (i.e., the $L_{2,1}$ norm regularization) for learning sparsity~\cite{Wen2017LearningIS}, the lasso based methods are shown to be insufficient in inducing sparsity for large scale non-convex problems such as the training of DNNs~\cite{DBLP:conf/cvpr/SrinivasSB17,DBLP:journals/corr/CollinsK14}. In contrast,  the expected $L_0$ minimization  closely resembles spike-and-slab priors \cite{mitchell1988bayesian,ZheXQY15,XuZQY16} used in Bayesian variable selection \cite{DBLP:conf/cvpr/SrinivasSB17,Louizos2017LearningSN}.
The spike-and-slab priors can induce high sparsity and encourage large values at the same time due to the richer parameterization of these priors, while LASSO shrinks all parameters until lots of them are close to zero.
And the $L_0$-norm regularization explicitly penalizes parameters for being different than zero with no other restrictions.
Hence compared with Intrinsic Sparse Structures (ISS) via Lasso proposed by~\cite{Wen2017LearningIS}, our neuron selection via the $L_0$ norm regularization can achieve higher adequate sparsity in RNNs. 

In this paper, we propose a new type of method to prune individual neurons of RNNs. Our key contribution is that we introduce binary gates on recurrent and input units such that sparse masks for the weight matrix can be generated, allowing for effective neuron selection under sparsity constraint. For the first work of neuron selection in RNNs, we attempt to employ the smoothed mechanism for the $L_0$ regularized objective proposed in~\cite{Louizos2017LearningSN}, motivated from~\cite{DBLP:conf/cvpr/SrinivasSB17}. 
 
We evaluate our structured pruning method on two tasks, i.e., language modeling and machine reading comprehension. For example, in the case of language modeling of the word level on the Penn Treebank dataset, our method achieves the state-of-the-art results, i.e., the model size is reduced by more than 10 times, and the inference of the resulted sparse model is nearly 20 times faster than that of the original model.
We also achieve encouraging results for the recurrent highway networks~\cite{zilly2017recurrent} on language modeling and BiDAF model~\cite{seo2016bidirectional} on machine reading comprehension.

\section{Related Work}
Despite model compression has achieved impressive success in DNNs(e.g., CNNs)~\cite{Han2016DeepCC,MOHAMMED201769,AYINDE2019}, it is difficult to directly apply this technology of compressing DNNs to the compression of RNNs due to the recurrent structure in RNNs. There are some recent efforts on the compression of RNNs. Generally, the compression techniques on RNNs can be categorized into the following types:  pruning ~\cite{narang2017exploring, narang2017block,Wen2017LearningIS},
low-rank matrix/tensor factorization~\cite{DBLP:conf/icassp/PrabhavalkarABM16,YeWLCZCX18,zilly2017recurrent} and quantization~\cite{DBLP:journals/corr/HubaraCSEB16,DBLP:conf/nips/WangXDLWX18}. \cite{DBLP:journals/spl/WangLW18} introduces several strategies including gate activation sparsity, top-k pruning schemes and mixed quantization schemes to compress LSTMs. Our work lies in the branch of pruning. 

Pruning approaches can be further divided into non-structured pruning and structured pruning. 
For non-structured pruning, elements of the weight matrices can be removed based on some criteria. 
For example, ~\cite{narang2017exploring} presents a magnitude-based pruning approach for RNNs, i.e., at every iteration, the top-k elements of the weights are set as 0. 
While such an approach can achieve over 90\% sparsity in RNNs of Deep Speech 2 model with a minor decrease of accuracy, the obtained non-structured sparse matrices cannot efficiently accelerate the computation in modern computing platforms due to the irregular memory access. 
To improve this, ~\cite{narang2017block} further proposed block-structured pruning in RNNs via the group lasso regularization.
It extends the approach in ~\cite{narang2017exploring} to prune blocks of a matrix instead of individual weights. 
~\cite{Wen2017LearningIS} also proposed Intrinsic Structured Sparsity (ISS) for LSTMs by collectively removing the columns and rows for the weight matrices. 
ISS reduces the sizes of basic structures within LSTM units and is more hardware-friendly for acceleration compared with block structure in ~\cite{narang2017block}. 
Similar to~\cite{narang2017block}, ISS also relies on group lasso for sparsity learning.
Nevertheless, lasso regularization is shown to be insufficient for large non-convex problems, e.g., the training of DNNs ~\cite{DBLP:journals/corr/CollinsK14}. 
To alleviate this challenge in~\cite{Wen2017LearningIS,narang2017block}, we instead present a structured sparsity learning method through $L_0$ regularization, which not only reduces the size of basic structures of LSTMs but also achieves higher sparsity for non-convex RNN training in a tractable way. 

\section{Structured pruning of LSTMs through neuron selection }\label{section:sslm}

Without loss of generality, we focus on the compression of LSTMs~\cite{hochreiter1997long}, a common variant of RNN that learns long-term dependencies. Note that our method can be readily applied to the compression of GRUs and vanilla RNNs. Before presenting the proposed sparsification methods, we first introduce the LSTM network.
\begin{eqnarray}
  i_t &= &\sigma(W^ix_t+U^ih_{t-1}+b_i), \nonumber\\
  f_t &=&\sigma(W^fx_t+U^fh_{t-1}+b_f), \nonumber\\
  o_t &= &\sigma(W^ox_t+U^oh_{t-1}+b_o),\nonumber\\
  u_t & =& \tanh(W^ux_t+U^uh_{t-1}+b_u),\\
  c_t &=& i_t\odot u_t +f_t\odot c_{t-1},\nonumber\\
  h_t &=& o_t \odot \tanh(c_t),\nonumber
  \end{eqnarray}
where $\sigma(\cdot)$ is the sigmoid function, $\odot$ denotes the element-wise multiplication and $\tanh(\cdot)$ is the hyperbolic tangent function. 
$x_t$ denotes the input vector at the time-step $t$, $h_t$ denotes the current hidden state, and $c_t$ denotes the long-term memory cell state. $i_t$, $f_t$ and $o_t$ correspond to the input gate, the forget gate and the output gate, respectively. 
For notation simplicity, we let $W=\{W^i,W^f,W^o,W^c\}$ be the input-to-hidden weight matrices, and $U=\{U^i,U^f,U^o,U^c\}$ be the hidden-to-hidden weight matrices.
\begin{figure*} \centering    
\subfigure[stacked LSTMs]{
 \label{fig:a}     
\includegraphics[width=0.7\textwidth,height=0.2\textheight]{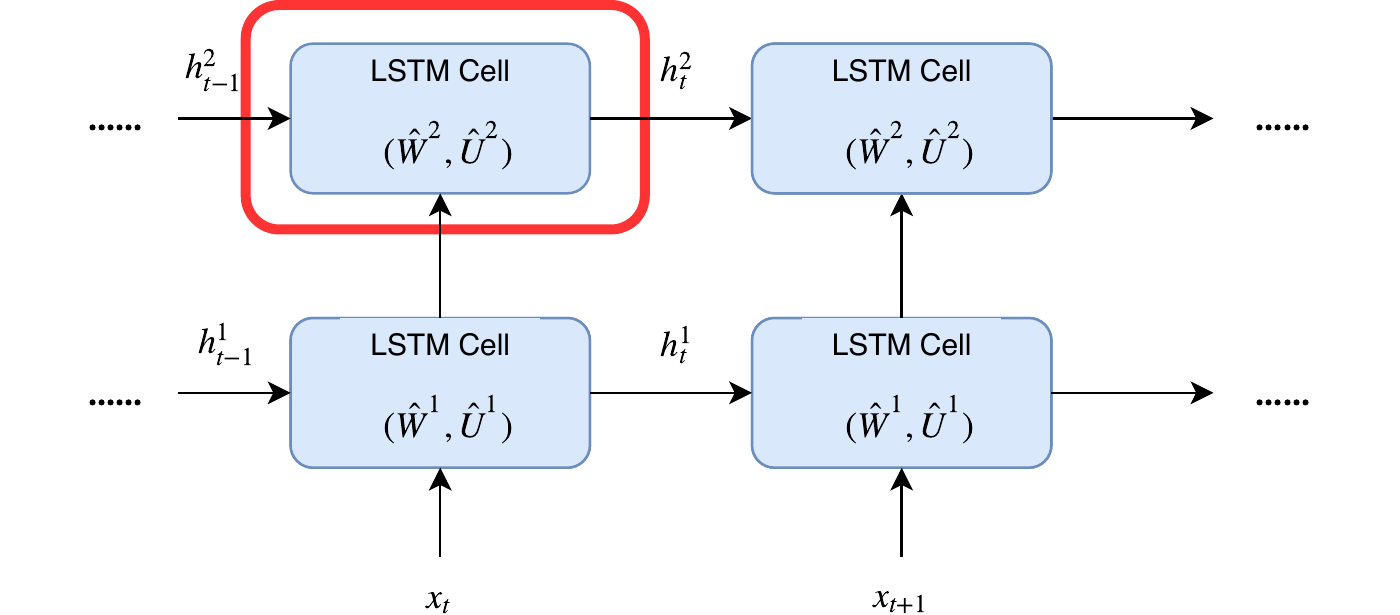}  
}     

\subfigure[input to hidden state] { 
\label{fig:b}     
\includegraphics[width=0.4\textwidth,height=0.2\textheight]{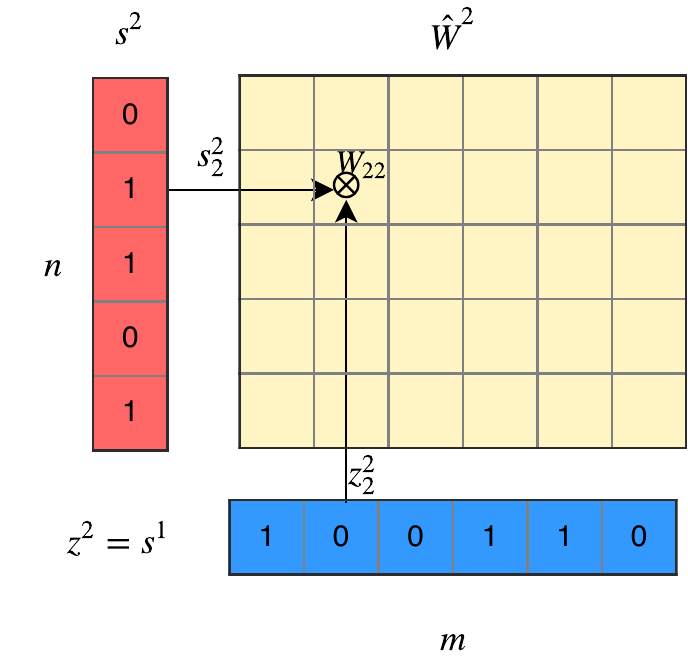}     
}    
\subfigure[hidden  to hidden state] { 
\label{fig:c}     
\includegraphics[width=0.4\textwidth,height=0.2\textheight]{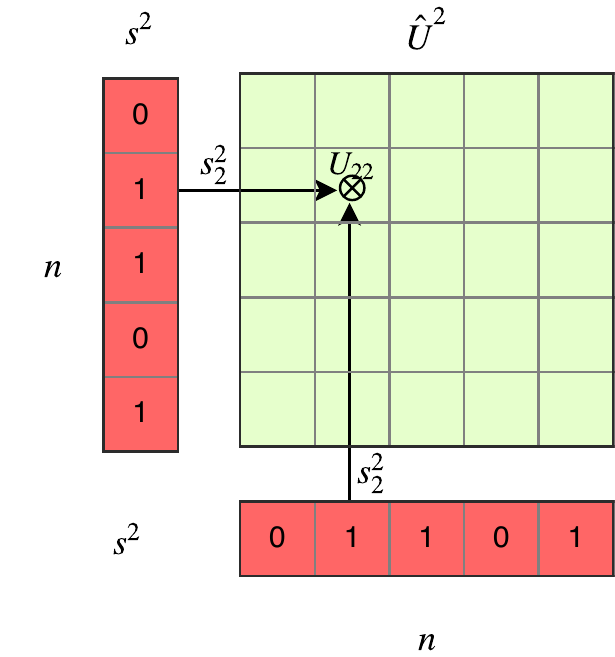}     
}   
\caption{ Illustration on the proposed gate mechanism. (a) shows an illustration of stacked LSTMs with $\hat{W}$ and $\hat{U}$, where the parameters in the red box are illustrated in (b) and (c); (b) shows the gate $z^{2}$ and $s^2$ controlling the input-to-hidden matrix $\hat{W}^2$ activation of the input neurons, where $z^2=s^1$ since the output of $W^1$ is the input to $W^2$; and (c) shows the gate $s^{2}$ controlling the hidden-to-hidden weight matrix $U^{2}$.}
\label{fig:gates}     
\end{figure*}
\subsection{ Neuron Selection via Binary Gates}\label{sec_str_s}
To prune individual neurons in LSTMs, which can simultaneously reduce the size, FLOPs of LSTMs, we introduce neuron selection mechanisms into the design of LSTMs. 
As illustrated in Figure~\ref{fig:gates} (b) and (c), we introduce two auxiliary sets of binary "gate" or "switch" variables: one set of variables (denoted by $z=\{z_i\}$) controls the presence of the input neuron $i$, where $z_i\in\{0,1\}$ and $|z|=m$ is the number of input neurons the other set of binary gate switch variables (denoted by $s={\{s_j\}}$) controls the presence of the hidden neuron $j$, where $s_j\in\{0,1\}$ and $|s|=n$ is the number of hidden neurons. In this way, $w_{ij}$ controls the strength of the link from the input neuron $i$ to the hidden neuron $j$, while $z_i$ and $s_j$ control the presence of neurons, and the mask on $w_{ij}$ can be calculated by $z_i\times s_j$. 
In particular, such a gating mechanism can induce structured sparsity on the weight matrices. If $z_i=0$, all hidden neurons connected to $i$ will be switched off meaning that the $i$-th column of $W$ will be all zeros; and $s_j=0$ will turn off each row of $W$. 
Therefore, by sharing the binary masks across all the gates, we can obtain structured sparsity on the weight matrices.
For convenience, we re-parameterize the original parameter matrices $W$ and $U$  to $\hat{W}$ and $\hat{U}$ as follows,
\begin{align}\label{equ_mask}
\hat{W} = W\odot (z s^{\top}),~~~~~~~ \hat{U} = U\odot (s s^{\top}),
\end{align}
where $\odot$ denotes the element-wise product operation.
In this way, $w_{ij}$ controls the strength of the link from the input neuron $i$ to the hidden neuron $j$, while $z_i$ and $s_j$ control the presence of neurons. 
Furthermore, our proposed switch mechanism can be directly applied to stacked LSTMs. As illustrated in  Figure~\ref{fig:a}, since the output of the $l$-th layer is the input of the $l+1$-th layer, we have $z^{l+1}=s^{l}$. 

To model the uncertainty of each random ``gate'' variable $z_i$ and $s_j$, we let $z_i \sim\mathrm{Bern}(\pi_{z_i})$ and $s_j\sim \mathrm{Bern}(\pi_{s_j})$, where $\pi_{z_i}$ and $\pi_{s_j}$ are the parameters of the Bernoulli distributions, and denote the probability of random variable $z_i$ and $s_j$ taking value 1,  respectively. 
We can optimize this objective function with $\pi_{z}$ and $\pi_{s}$ by minimizing the $L_0$ norm of the weight matrix to achieve neuron selection. 
The $L_0$ norm regularization can explicitly penalize no-zero parameters of models without further restrictions~\cite{Louizos2017LearningSN}, and it demonstrates superior advantages over the $L_1$ norm regularization in the sparse learning. A naive idea in learning sparsity is to directly utilize the $L_0$-norm regularization on the weights of LSTMs, which leads to the following objective function:
\begin{align}
 \label{general_obj_loss}
 &\mathcal{L}(W,U) =E_{D} (W,U)+\lambda_1\|W\|_0 + \lambda_2\|U\|_0, \nonumber \\
 &W^*,U^* = \arg\min \limits_{W,U} \mathcal{L}(W,U). 
\end{align}
Here $\|W\|_0=\sum_{i=1}^{m}\sum_{j=1}^{n} \mathbb{I}(w_{ij}\ne 0)$ denotes the $L_0$-norm, and  $\|U\|_0$ follows a similar pattern. $\mathbb{I}(.)$ is the indicator function, and $m$ and $n$ denote the number of input units and hidden units,  respectively.  ${E}_{D}(W,U)$ represents the loss on the dataset $D$, and $\lambda_1$ and $\lambda_2$ are penalty parameters for the sparsity regularization. 

Neuron selection via binary random  variables  generates  masks  for  the  weight matrix. With the gating mechanism, the $L_0$ norm in Equation~\ref{general_obj_loss} can be further specified as $\|\hat{W}\|_0 = \sum_{i}\sum_{j} z_i\times s_j$ and $\|\hat{U}\|_0 = \sum_{j_1}\sum_{j_2} s_{j_1}\times s_{j_2}$. Hence we can seek to penalize the number of parameters appearing in LSTM on average. The expectation of Equation~ (\ref{general_obj_loss}) over the auxiliary masks is reformulated  as follows,
\begin{align}
\label{equ_expectation}
&\mathcal{L}(W, U, \pi_z,\pi_s)=\mathbb{E}_{\Phi(z|\pi_z)\Phi(s|\pi_s)}[\mathbb{E}_D(W,U,z,s)]\nonumber\\
&+\lambda_{1}\sum_{i=1}\sum_{j=1}\pi_{z_{i}} \pi_{s_{j}} + \lambda_{2}\mathop{\sum_{i=1}\sum_{j=1}}^{i\ne j}\pi_{s_{i}}\pi_{s_{j}}+\lambda_{2}\sum_{j=1}\pi_{s_{j}},\\
& (W^* ,U^*, \pi_z^*,\pi_s^*)=\mathop{\arg\min}_{W ,U,\pi_z,\pi_s}\mathcal{L}(W, U,\pi_z,\pi_s).\nonumber
\end{align}
The item $s_j\times s_j$ in $\|\hat{U}\|_0 = \sum_{j_1}\sum_{j_2} s_{j_1}\times s_{j_2}$ denotes the mask of weight parameters $U_{jj}$. The term $s_j\times s_j$  only depends on the presence of the $j$-th recurrent units. Hence the expectation of $s_j\times s_j$ is $\pi_j$. Note that to avoid quadratic terms on $s_j$, we write $\|\hat{U}\|_0$ as the summation of two terms as shown in Equation~(\ref{equ_expectation}).
Since the binary gates are shared across layers in stacked RNNs, it may result in an extremely unbalanced sparse structure in different layers to penalize the gate variables of different layers with the same penalty factor. For this reason, we specify independent regularization for different layers.

\subsection{Optimization} 

It is intractable to learn sparse parametric models by minimizing the $L_0$ norm based on gradient optimization. The optimization of the objective in Equation~(\ref{equ_expectation}) is problematic due to the discrete nature of $z$ and $s$.
In principle, the REINFORCE estimator~\cite{williams1992simple} can be used to compute the gradients, but it suffers from high variance and slow convergence. 
On the other hand, the straight-through estimator (STE)~\cite{bengio2013estimating} can also be used, however, the mismatch of the parameters between the forward and backward pass in the optimization leads to biased gradients and updates.

A more appealing method is to continuously relax discrete random binary variables by a hard-sigmoid rectification of continuous random variable with a distribution~\cite{Louizos2017LearningSN}, as follows:
\begin{gather}
    z = \min(\mathbf{1},\max(\mathbf{0},\tau_z)), \quad \tau_z \sim q(\tau_z|\phi_z),
\end{gather}
where the $q(\tau_z|\phi_{z})$ corresponds to the continue distribution with parameter $\phi_{z}$.  
Then the probability of the  discrete random binary variables being non-zero is computed by the cumulative distribution function(CDF) $Q(.)$ of $\mathbf{s}$, as follow:
\begin{gather}
\Phi_{z}(z\neq 0|\phi_z)=1-Q(\tau_z \le 0|\phi_z).
\end{gather}
Then it can not only enable gradient-based optimization of a generic loss by smoothing the binary gate variables $\mathbf{z}$,
but also allow the variables $z$ to be exactly zero.
We can write the continuous distribution $q(\tau_z|\phi_{z})=q(f(\phi_{\tau_z},\epsilon))$ by using the the reparameterization trick~\cite{kingma2013auto,Rezende2014StochasticBA}, where $f(.)$
is a deterministic and differentiable function and $\epsilon$ denotes the uniform or Gaussian free noise. Since $\epsilon$ is independent of the parameters of models, we can directly take the gradient of the optimization target over the parameters of the distributions of random variables. 

Similarly, we can apply the same procedure to smooth gate variables $s$ with $\phi_s$ which denotes the parameters of the correlative continuous distribution. Armed with the above approach, we can generate $z$ and $s$ via the differentiable transformation in Equation~(\ref{hard}), and therefore shift the optimization over $\pi_z$ and $\pi_s$ in Equation~(\ref{equ_expectation}) to $\phi_z$ and $\phi_s$ in the hard concrete distributions. The original objective function in Equation~(\ref{equ_expectation}) can be reformulated as
\begin{align}
\label{final_loss}
&\tilde{\mathcal{L}}({W},{U},{\phi_z},\phi_s)\nonumber=\mathbb{E}_{\Phi(z|\phi_z)\Phi(s|\phi_s)}[\mathbb{E}_D({W},{U},z,s)] \nonumber\\
&+\lambda_{1}\sum_{i=1}\sum_{j=1}(\Phi_{z_{i}}(z_i\neq 0|\phi_{{z}_i}))(\Phi_{s_{j}}(s_j\neq 0|\phi_{{s}_j}))\nonumber\\
&+\lambda_{2}\mathop{\sum_{i=1}\sum_{j=1}}^{i\ne j}(\Phi_{s_{i}}(s_i\neq 0|\phi_{{s}_i}))(\Phi_{s_{j}}(s_j\neq 0|\phi_{{s}_j}))\nonumber\\
&+\lambda_{2}\sum_{i=1}\Phi_{s_{i}}(s_i\neq 0|\phi_{{s}_i})).\nonumber\\
& (W^*, U^*, \phi_z^*, \phi_s^*) = \mathop{\arg\min}_{W ,U,\pi_z, \pi_s}\tilde{\mathcal{L}}(W,U,\Phi_z,\Phi_s).
\end{align}

As shown in~\cite{Louizos2017LearningSN}, an efficient choice of the smoothing continuous distribution is as following; the binary concrete distribution~\cite{maddison2016concrete,DBLP:conf/iclr/JangGP17} $q(\hat{\tau})$ with the parameters $\log \alpha$ and $\beta$, where $\log \alpha$ and $\beta$ are the location and temperature parameters respectively, is stretched from the (0,1) interval to the 
($\zeta$,$\gamma$) interval,with $\zeta_z < 0$ and $\gamma_z > 1$. Then we apply a hard-sigmoid on its random samples.  More specific, the  procedure to smooth gate variables $s$ is as following:
\begin{align}
\label{hard}
u & \sim \mathrm{Uniform}(0,1), \nonumber\\
\hat{\tau}&= \mathrm{\sigma}((\log u-log(1-u)+\log\alpha_z)/ \beta_z),\\
\tau&=\hat{\tau}(\zeta_z-\gamma_z)+\gamma_z,\nonumber\\
z&=\min(1, \max(0,\tau )), \nonumber
\end{align}
where $\sigma$ is the sigmoid function as introduced before.
The probability of the $z$ being non-zero can be computed by the cumulative density function $\Phi(\cdot)$ of $z$ as follows:
\begin{align}
\Phi_{z}(z\neq 0|\phi_z) = \mathrm{\sigma}(\log\alpha_z-\beta_z \log\frac{-\gamma_z}{\zeta_z}),  
\end{align}
where $\phi_z=\{\alpha_z, \beta_z, \zeta_z, \gamma_z\}$ denotes the parameters of the hard concrete distribution. Similarly, we can apply the same procedure to smooth gate variables $s$.

During testing, we use the following estimator for the final $z$ and $s$ under a hard concrete smoothness:
\begin{align}
z &=\min(1,\max(0,\mathrm{\sigma}(\log\alpha_z)(\zeta_z-\gamma_z)+\gamma_z)),\\
s &=\min(1,\max(0,\mathrm{\sigma}(\log\alpha_s)(\zeta_s-\gamma_s)+\gamma_s)).
\end{align}

\section{Experiments}
To compare with Intrinsic Sparse Structures (ISS) via Lasso proposed by~\cite{Wen2017LearningIS}, we also evaluate our structured sparsity learning method with $L_{0}$ regularization on language modeling and machine reading tasks. In the case of language modeling, we seek to sparsify a stacked LSTM model \cite{zaremba2014recurrent} and the state-of-the-art Recurrent Highway Networks (RHNs)~\cite{zilly2017recurrent}
\baicomment{with a large hidden size of 1500 
on word-level}
on the Penn Treebank~(PTB) dataset~\cite{marcus1993building}. 
For the task of machine reading comprehension, we choose the Bi-Directional Attention Flow Model (BiDAF) \cite{seo2016bidirectional} with a small hidden size of 100 on the SQuAD dataset \cite{rajpurkar2016squad}. While our structured $L_0$ norm imposes no shrinkage on the remaining components, the learned model could be over-fitted if weight decay is not assigned. Consequently, we follow a similar pattern in~\cite{Louizos2017LearningSN} to impose $L_2$ regularization on model parameters. For the setting of the hard concrete distribution, we follow the same pattern in \cite{Louizos2017LearningSN} for all experiments, i.e., $\zeta, \gamma, \beta$ are taken as the hyper-parameters.  For $\log\alpha$, it is updated by back-propagation of the network and initialized by samples from $\mathcal{N}(1, 0.1)$.

\subsection{Language Modeling}
For language modeling, we evaluate two models: stacked LSTMs and recurrent highway neural networks. Both models are trained from scratch. We use the word level PTB dataset for language modeling, which consists of 929k training words, 73k validation words and 82k test words with 10,000 unique words in its vocabulary. 

\subsubsection{Stacked LSTMs}

\textbf{Baselines}. We compare our proposed method against two baselines, the vanilla two-layer stacked LSTM used in \cite{zaremba2014recurrent} and the ISS method~\cite{Wen2017LearningIS}, which is the state-of-the-art method in RNN compression. The dropout keep ratio is $0.35$ for the vanilla model. The vocabulary size, embedding size and hidden size of the stacked LSTMs are set as $10,000$, $1,500$ and $1,500$, respectively, which is consistent with the settings in ~\cite{Wen2017LearningIS}. The results of ISS are taken from the original paper.

\textbf{Hyper-parameters Setting}. 
We use NT-ASGD~\cite{merity2017regularizing} for training with an initial learning rate equal to $20.0$ and the gradient clipping set to $0.25$.
Similar to ISS~\cite{Wen2017LearningIS}, we increase the dropout keep ratio to 0.65 due to the intrinsically structured sparsity in the network.
We use the default initialization strategy provided in PyTroch\footnote{https://pytorch.org/} for the input and output word embedding as well as the parameters of the LSTM. 


\newcommand{\tabincell}[2]{\begin{tabular}{@{}#1@{}}#2\end{tabular}}
\begin{table*}[t]
        \caption{Learning structured sparsity from scratch in stacked LSTMs.}
		\label{tab:wLSTM}
		\centering 
		\resizebox{\textwidth}{15mm}{
		\begin{tabular}{lccccccc}  
			\toprule
			Method      & \tabincell{c}{Droupout \\keep ratio }&\tabincell{c}{ Perplexity\\(validate, test) }&\tabincell{c}{LSTMs\\(input,L1,L2)} & Size & Time (ms)  & Speedup & \tabincell{c}{ Multi-add \\reduction }\\
			\midrule
			Vanilla Model & 0.35  & (82.57,78.57)&(1500, 1500, 1500)  &66.0M & 365.92 $\pm$ 10.3 &1.00$\times$&1.00$\times$\\
			\midrule
			ISS & 0.60  & (82.59, 78.65)&(1500, 373, 315)  &21.8M & 39.88 $\pm$ 0.7 &9.17$\times$&7.48$\times$\\
			our method    & 0.65  & (\textbf{81.62}, \textbf{78.08})& (\textbf{251}, \textbf{296}, \textbf{247}) & \textbf{6.16M}  & \textbf{18.87 $\pm$ 0.3}& \textbf{19.39$\times$} &\textbf{13.95$\times$} \\
			\bottomrule
		\end{tabular}}
\end{table*} 

\begin{figure*}

\setlength{\tabcolsep}{12pt} 
\renewcommand{\arraystretch}{1} 
\begin{tabular}{c}
\includegraphics[width=.45\textwidth]{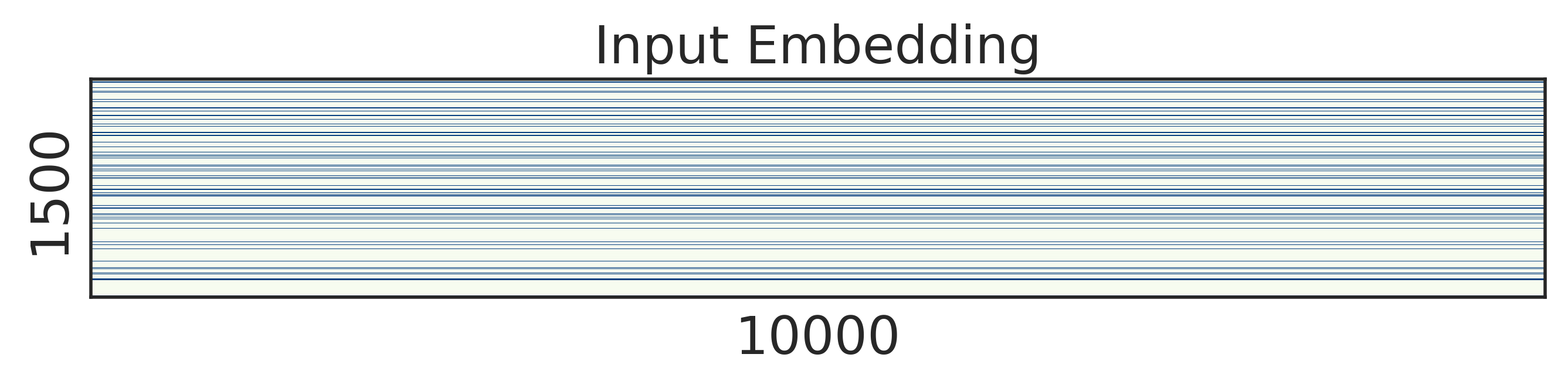} 
\hspace{16.0em}\llap{
\raisebox{0.0em}{\includegraphics[width=.45\textwidth]{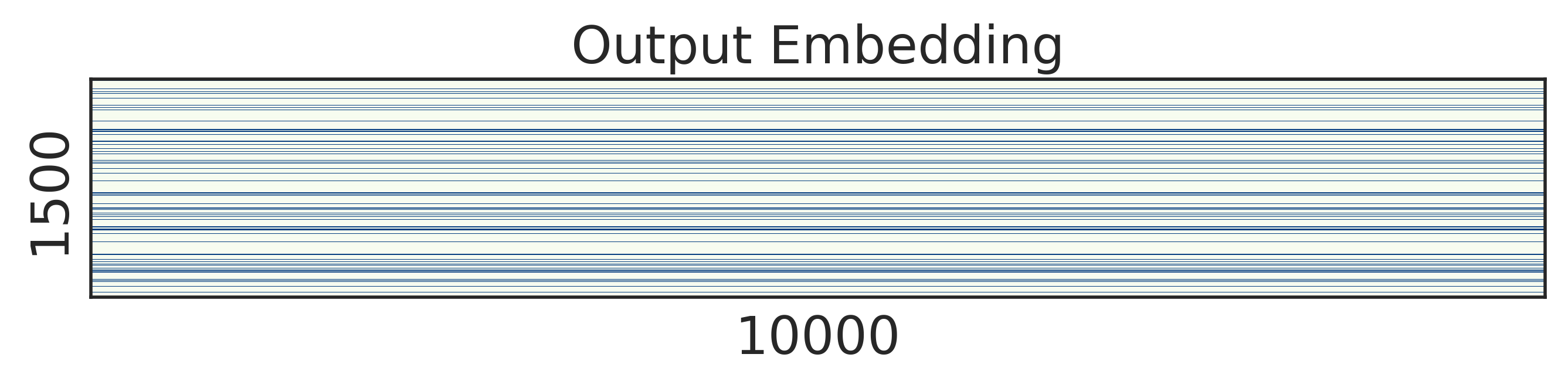}}}  \\
\includegraphics[width=.45\textwidth]{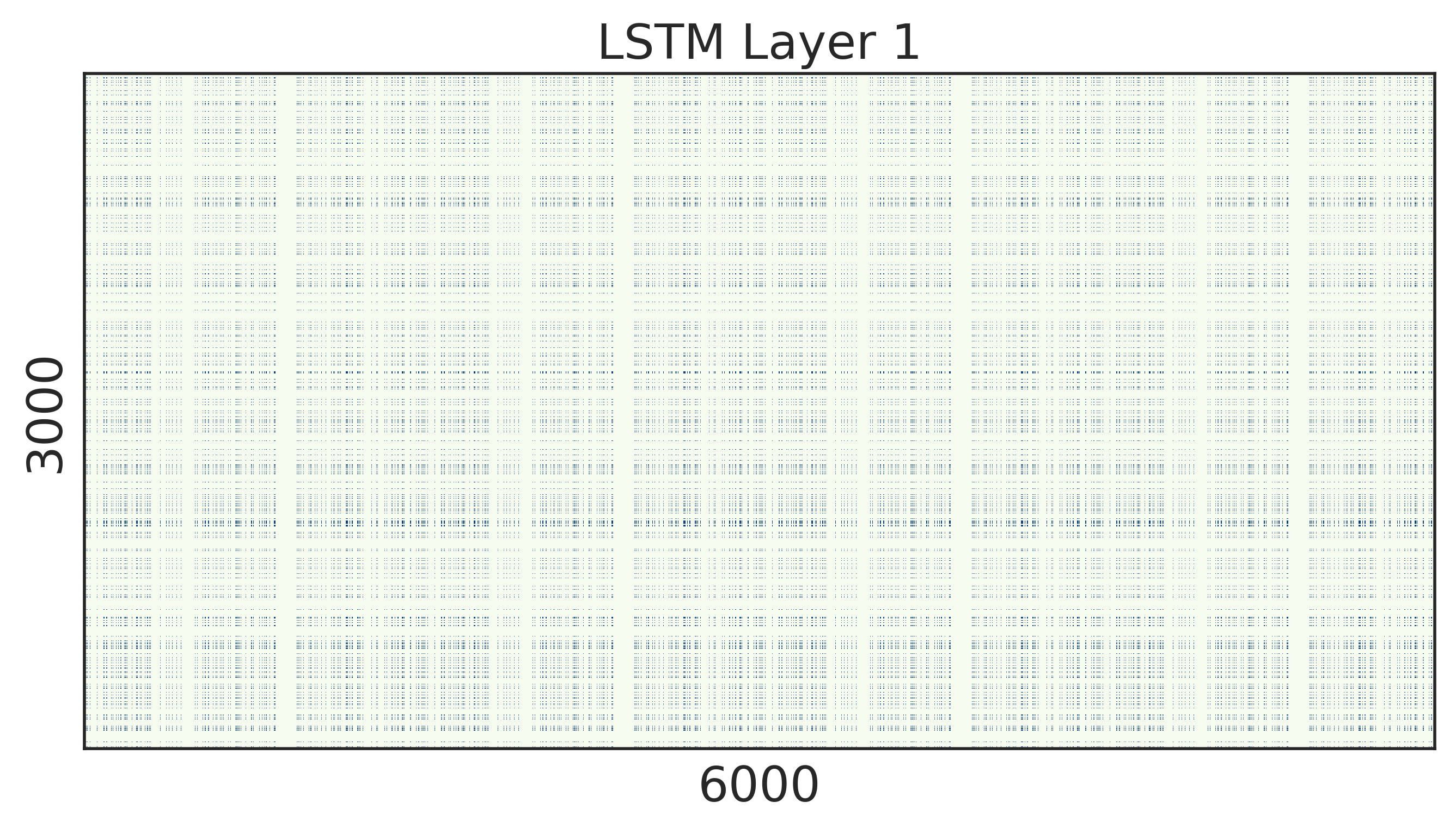} 
\hspace{16.0em}\llap{
\raisebox{0.0em}{\includegraphics[width=.45\textwidth]{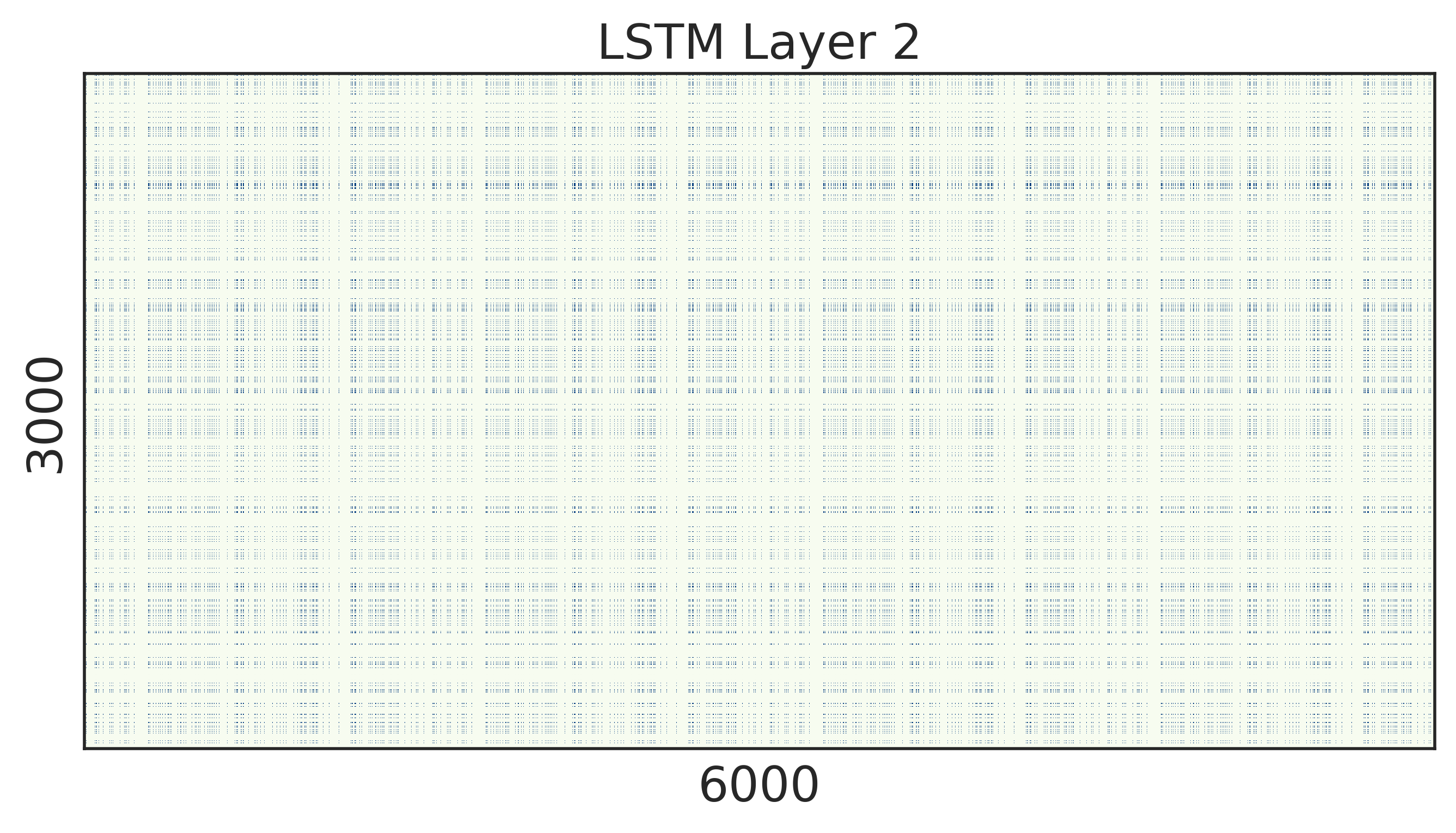}}}
\end{tabular}
\caption{Illustration of word embedding and weight matrices in the two-layer LSTM. The first row presents the embedding layer for the input and output respectively. The second row shows the weight matrices for the first and second layers. Note that we have concatenated $\{W^i, W^f, W^u, W^o\}$ and $\{U^i, U^f, U^u, U^o\}$ into a single matrix with shape $3,000 \times 6,000$. The blue dots are nonzero weights and the rest ones are structurally pruned to zeros.
} 
\label{fig:ss_lstm}
\end{figure*}

 \begin{table*}[t]
\caption{Ablative study the penalty parameters of stacked LSTMs.}
\label{tab:ablative}
\centering
\begin{threeparttable}
\begin{tabular}{ccc}  
\toprule
\tabincell{c}{The penalty scale \\  $\lambda_1^{(1)}$,$\lambda_2^{(1)}$,$\lambda_1^{(2)}$ and $\lambda_2^{(2)}$} & \tabincell{c}{Perplexity\\(validate, test)} &\tabincell{c}{the dimension of \\ input, 1st, 2nd} \\
\midrule (1, 1, 1, 1)& (81.73, 78.33)    &(1020, 215, 250)  \\
\midrule
(2, 1, 1, 1)    & (82.87, 79.25)    &(635, 146, 247) \\
\midrule
(3, 1, 1, 1)  & (85.49, 81.82)    &(438, 110, 244)\\
\midrule
(4, 1, 1, 1)  & (87.37, 84.22)    &(333, 88, 242)\\
\midrule
(5, 1, 1, 1)  & (90.83, 84.22)    &(273, 70, 237)\\
\bottomrule
\end{tabular}
\begin{tablenotes}
\footnotesize
\item[*]  We use $\lambda_1^{(1)}$,$\lambda_2^{(1)}$,$\lambda_1^{(2)}$ and $\lambda_2^{(2)}$ to denote the times to  $0.08/N$, e.g., $ \lambda_1^{(1)} \times 0.08/N$ denotes the penalty parameter to the input neuron the first layer. 
\end{tablenotes}
\end{threeparttable}
\end{table*}
\textbf{Results}. We show the results of stacked LSTMs in Table \ref{tab:wLSTM}. It can be observed that our method finds the most compact structure of the model, i.e. the numbers of the first and second hidden units are reduced from 1500 to 296 and 247 respectively, both of which are significantly smaller than the vanilla stacked model and the ISS method. Besides, the dimension of word embedding vectors also decreases from 1500 to 251. 
Overall, our method reduces the model size form $66.0M$ to $6.16M$, which is more than 10$\times$ reduction comparing to the vanilla model. Theoretically, the computation is reduced by nearly 14$\times$ in terms of multi-add operations.
Compared with ISS in \cite{Wen2017LearningIS}, our structured sparsity learning method reduces the multi-add computation further by 1.86x and the model size further by $15.64M$.
The results indicate that our structured $L_0$ regularization can indeed sufficiently sparse the model.
Additionally, despite the model size being sharply shrunk, our method still achieves the lowest perplexity on the PTB dataset. The excellent performance of our method can be explained by the superiority of our structured $L_0$ regularization since it poses no penalization over the remained parameters, while for ISS, the group lasso method penalizes the norms of all groups collectively, and thereon could affect the model capacity.

In order to evaluate the practical speedup of the learned structures of our proposed method and all the baseline, we measure the speedup of inference on CPU\footnote{Intel CPU E5-2630 v4 @ 2.20GHz processor with a total of 40 cores.} using TensorFlow with Intel MKL library\footnote{https://www.tensorflow.org/guide/performance/overview.}. The time is measured with 10 batch size and 30 unrolled steps, and the result is averaged from $1,000$ times of inference with standard deviation reported. It can be seen from Table~\ref{tab:wLSTM} that the practical inference time is 19.4$\times$ faster than the original model. The actual speedup is even higher than the theoretical result, and we conjecture that this could be due to some basic optimization in the MKL library.


 To look into the learned structured sparsity, we further visualize the embedding and weights of the stacked LSTMs in Figure ~\ref{fig:ss_lstm} after training 200 epochs. We can see that after structured $L_0$ regularization, the size of word embedding is highly reduced, and similarly, most rows and columns of the weight matrices are pruned away. Therefore, such matrices can be re-arranged to a small and compact structure, leading to practical speed up during inference.

\textbf{Ablative Study}. 
 We used a two-layer stacked LSTM to verify the sensitivity of the penalty parameters in Equation~(\ref{general_obj_loss}). Experimentally, we find that setting all the penalty values to $0.08/N$ (where $N$ denotes the number of training data) can lead to good performance. For convenience, we use $\lambda_1^{(1)}$,$\lambda_2^{(1)}$,$\lambda_1^{(2)}$ and $\lambda_2^{(2)}$ to denote the times to  $0.08/N$, e.g., $ \lambda_1^{(1)} \times 0.08/N$ denotes the penalty parameter to the input neuron the first layer. Here the super-index denotes the corresponding layer.  Initially, we set all the parameters to the same value (i.e., $0.08/N$). We find that the numbers of the first-layer and second-layer hidden units are reduced from 1500 to 215 and 250 respectively, while the size of the input only decreases from 1500 to 1020.

 Since the first input layer is usually much larger than the hidden layers, we increase the penalty value of $\lambda_1^{(1)}$ to drop out more input neurons and to seek much smaller networks. From Table 2, we can find that with a larger value of $\lambda_1^{(1)}$, the sparser input layer can be obtained while with a significant increase in the perplexity.

\subsubsection{Recurrent Highway Networks}
Our method can be further extended to Recurrent Highway Networks~(RHNs)~\cite{zilly2017recurrent}. 
Due to the structure of RHNs, we only introduce the binary mask $z$ and set its regularization as $\lambda$. During training, we impose $L_0$ regularization over $z$ so as to learn structured sparsity.
\baicomment{Following the same idea of structured sparsity to reduce the size of basic structures in LSTMs, we can identify sparse structure via binary gates $z$ to reduce RHNs width.  
The binary gate $z_i$ is introduced to represent whether the $i$th RHN width unit is present. 
The weight matrices of the H nonlinear transform, of the T and C gates in RHNs can be re-parametrized like Equation~(\ref{equ_mask}). Hence, we can learn structured sparsity  through $L_{0}$ Regularization. }

\textbf{Baselines}.
To evaluate our structured sparsity learning method, we choose "Variational RHN + WT" of \cite{zilly2017recurrent} as our baseline model. The number of units per layer is defined as the width of RHNs. It has depth 10 and width 830, with totally $23.5M$ parameters. The implementation of RHNs is available from its authors\footnote{https://github.com/jzilly/RecurrentHighwayNetworks}.
Aside from the vanilla RHNs, we also compare to the ISS method, and its results are taken from ~\cite{Wen2017LearningIS}.

\begin{table*}[t]
\caption{Learning structured sparsity from scratch in RHNs}
\label{tab:rhn}
\centering
\begin{tabular}{lccc}  
\toprule
Method      & Perplexity(validate, test) &RHN width &Parameter\\
\midrule RNHs~\cite{zilly2017recurrent} & (67.9, 65.4)    & 830  &23.5M \\
\midrule
ISS~\cite{Wen2017LearningIS}        & (68.1, 65.4)    &517  &11.1M\\
\midrule
our method  & (\textbf{68.2, 65.1})   & \textbf{389} & \textbf{7.2M}\\
\bottomrule
\end{tabular}
\end{table*}
\begin{figure}
\centering
{
\setlength{\tabcolsep}{12pt} 
\renewcommand{\arraystretch}{1} 
\begin{tabular}{c}
\raisebox{5.0em}{\rotatebox{90}{\small{ Perplexity(test)}}}
\includegraphics[width=.5\textwidth]{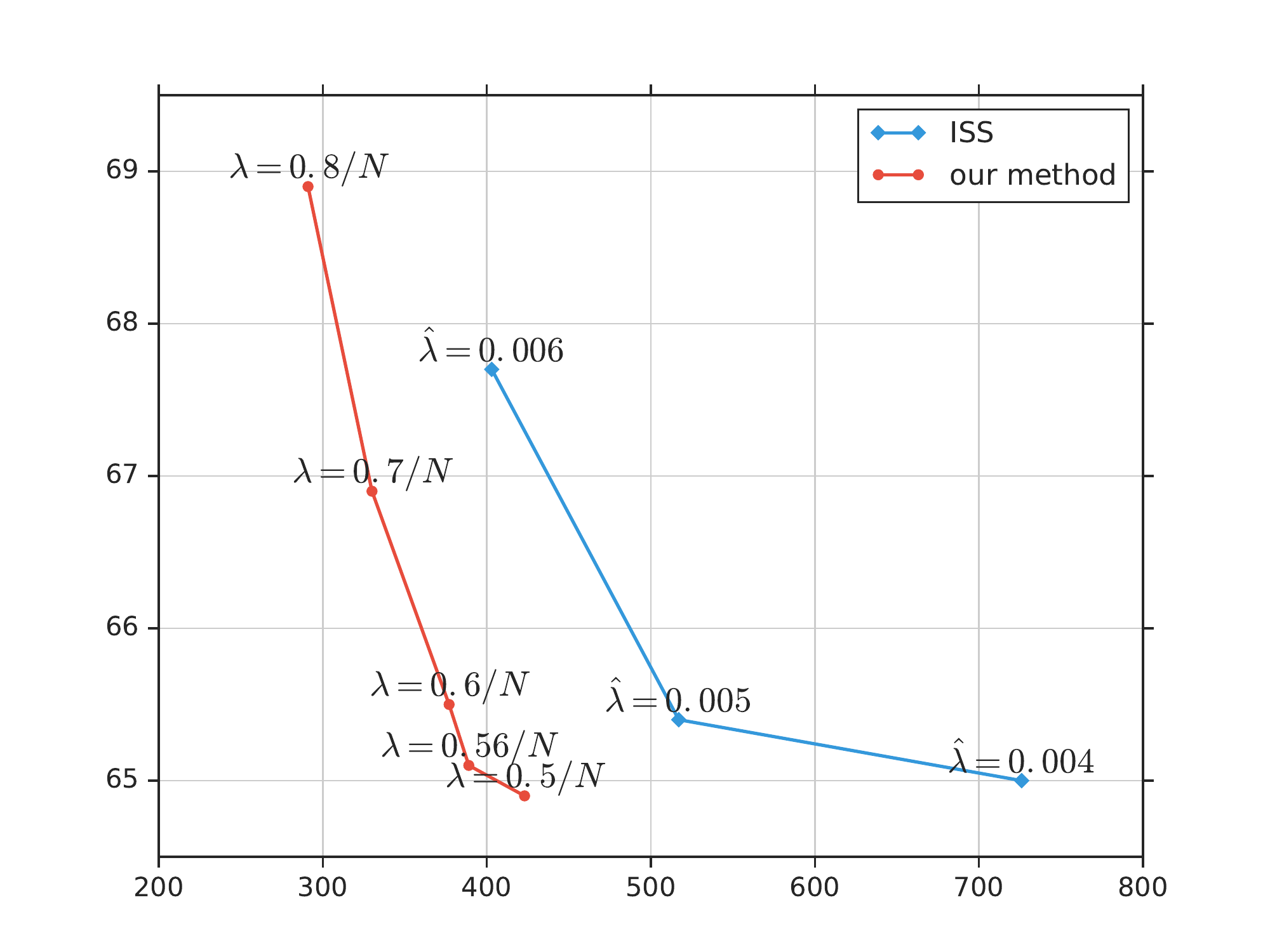}
\hspace{-1em} \\
\small{RHN width}
\end{tabular}}
\caption{Perplexity of our method and ISS under different sparsity levels with different regularization strength. For our method, we vary $\lambda$ in $\{0.5/N, 0.56/N, 0.6/N, 0.7/N, 0.8/N\}$(N denotes the size of the training data sample). In terms of ISS, we tune the group lasso regularization $\hat{\lambda}$ defined in its paper among $\{0.006, 0.005, 0.004\}$.
}
\label{fig:l0_res}
\end{figure}

\textbf{Hyper-parameters Setting}.
For our method, we use the same hyper-parameters as those in the baseline, except for that the parameters of models are initialized uniformly in $[-0.08,0.08]$, the dropout ratios are multiplied by $0.75$, and we divide the learning rate by a factor of $1.02$ at every epoch after it reaches $35$.

\textbf{Results}.
Table \ref{tab:rhn} shows the results obtained by the baseline, ISS and our method. Comparing to the vanilla RHN and ISS method, our structured sparsity approach can significantly achieve a more compact model with width 387 without losing perplexity. The parameter size also decreases to $7.1M$, which is about $69.8\%$ reduction.

We further investigate the trade-off between perplexity and sparsity of our method and ISS, and the plot is shown in Figure~\ref{fig:l0_res}.
It can be observed that our method can achieve a higher reduction of RHNs width than ISS at the same perplexity. With the same width, our method can achieve lower perplexity as well. This again demonstrates the superiority of our structured $L_0$ regularization over group lasso methods in inducing sparsity for recurrent structures.

\subsection{Machine Reading Comprehension}
 Machine Reading Comprehension (MRC) is one of the frontier tasks in the field of natural language processing.
 In MRC, the models answer a query about a given context paragraph, and Exact-Match (EM) and F1 scores are two major metrics for the evaluation (the higher the better).
 We use the benchmark Stanford Question Answering Dataset (SQuAD)~\cite{rajpurkar2016squad}, which consists of 100,000+ questions crowd-sourced on more than 500 Wikipedia articles. 
 
\textbf{Baselines}.
We use the BiDirectional Attention Flow Model (BiDAF)\footnote{We use the code from  https://github.com/allenai/bi-att-flow.} \cite{seo2016bidirectional} as the backbone model to evaluate our structured sparsity learning method. The BiDAF is composed of bidirectional LSTMs, and our structured $L_0$ regularization method can be readily applied. We focus on sparsifying the two layers of the bidirectional LSTM  denoted as ModFwd1, ModBwd1, ModFwd2, and ModBwd2, since they are shown to be most computationally expensive layers in~\cite{Wen2017LearningIS}.

\textbf{Hyper-parameter setting}.
We penalize different layers of the model in a similar pattern to the two-layer LSTMs used language modeling.
We also increase the dropout keep ratio to 0.9 as the structured sparsity itself can prevent over-fitting to some extent.
All the rest training schemes are the same as those in the baseline.

\begin{table*}
\caption{Comparison between the vanilla trained BiDAF, ISS method and our method. }
\label{tab:bitf}
\centering
\resizebox{\textwidth}{12mm}{
\begin{tabular}{lccccccccccc}  
\toprule
Method       & EM    & F1 &ModFwd1 &ModBwd1 & ModFwd2 &ModBwd2 &outFwd & outBwd& Weight\\
\midrule
Vanilla BiDAF & 67.98  & 77.85  &100  &100 &100 &100  &100 &100 & 2.69M \\
ISS~\cite{Wen2017LearningIS}         & 65.36  & 75.78  &\textbf{20}  &\textbf{33}  &40  &38   &\textbf{31}  &16   & \textbf{0.95M} \\
our method   & 65.67 & 75.69   &26   &\textbf{33}  &\textbf{36}  &\textbf{36}   &33  &\textbf{15}  &  0.96M\\
\bottomrule
\end{tabular}}
\end{table*}
\textbf{Results}.
Table \ref{tab:bitf} shows the EM, F1, the number of remaining components and model sizes obtained by the baseline, ISS and our method. 
As is mentioned in \cite{Wen2017LearningIS}, the scale of the vanilla BiDAF is compact enough on the SQuAD dataset, and it is thereon hard to reduce the hidden size of those LSTM layers in BiDAF without losing any EM/F1. 
Our structure sparsity method achieves competitive sparsity and performance comparing to the ISS method, both of which produce remarkably compressed models under acceptable degradation to the vanilla BiDAF. 
The result again demonstrates that our $L_0$ structured methods can be effectively used to discover the sparsity of recurrent neural networks.

\section{Conclusion} 
In this paper, we propose a novel structured sparsity learning method for recurrent neural networks. By introducing binary gates on neurons, we penalize weight matrices through $L_0$ regularization, reduce the sizes of the network parameters significantly and lead to practical speedup during inference. We also demonstrate the superiority of our relaxed $L_0$ regularization over the group lasso used in previous methods.
Our methods can be readily used in other recurrent structures such as Gated Recurrent Unit, and Recurrent Highway Networks. 

For future work, we plan to explore the sparsity constraints for neuron selection  for further reducing the number of model parameters, to exploit a more appealing unbiased, lower variance estimator(e.g., the unbiased ARMs-estimator recently proposed in~\cite{DBLP:conf/iclr/YinZ19}) for neuron selection. We also plan to combine neuron selection with quantization algorithms to further reduce model sizes and FlOPs of RNNs.

\section*{Acknowledgments}

This work was partially supported by NSF China(No. 61572111), a Fundamental Research Fund for the Central Universities of China (No.ZYGX2016Z003), and Startup Funding (No.G05QNQR004). 

\section*{References}
\bibliography{main}

\end{document}